\documentclass[letterpaper, 10 pt, journal, twoside]{IEEEtran}

\usepackage{graphics}
\usepackage{amsmath}
\usepackage{amssymb}
\usepackage{multicol}
\usepackage{graphicx}
\usepackage{layouts}
\usepackage{wrapfig}
\usepackage{caption}
\usepackage{graphicx,subcaption}
\usepackage{easyReview}
\usepackage{printlen}
\usepackage{bm}
\usepackage{siunitx}
\usepackage{hyperref}

\title{Learning Variable Impedance Control for Contact Sensitive Tasks}

\author{Miroslav Bogdanovic$^{1}$, Majid Khadiv$^{1}$ and Ludovic Righetti$^{1, 2}$%
\thanks{This work was supported by New York University, the European Union’s Horizon 2020 research and innovation program (grant agreement No 780684 and European Research Council’s grant No 637935) and a Google Faculty Research Award.}
\thanks{$^{1}$Max-Planck Institute  for  Intelligent  Systems,  T\"ubingen,  Germany
        {\tt\small \{mbogdanovic, mkhadiv, lrighetti\}@tue.mpg.de}}%
\thanks{$^{2}$Tandon  School  of  Engineering,  New  York  University,  USA} }

\begin{document}

\maketitle

\begin{abstract}

Reinforcement learning algorithms have shown great success in solving different problems ranging from playing video games to robotics.
However, they struggle to solve delicate robotic problems, especially those involving contact interactions.
Though in principle a policy directly outputting joint torques should be able to learn to perform these tasks, in practice we see that it has difficulty to robustly solve the problem without any given structure in the action space.
In this paper, we investigate how the choice of action space can give robust performance in presence of contact uncertainties.
We propose learning a policy giving as output impedance and desired position in joint space and compare the performance of that approach to torque and position control under different contact uncertainties.
Furthermore, we propose an additional reward term designed to regularize these variable impedance control policies, giving them interpretability and facilitating their transfer to real systems.
We present extensive experiments in simulation of both floating and fixed-base systems in tasks involving contact uncertainties, as well as results for running the learned policies on a real system (accompanying videos can be seen 
\href{https://youtu.be/AQuuQ-h4dBM}{here}).

\end{abstract}


\section{INTRODUCTION}

\IEEEPARstart{M}{any} interesting robotic applications necessitate complex physical interactions with the environment. 
During locomotion, intermittent contacts and force modulation enable the robot to keep balance and move forward. Multi-contact
interactions are also central to the efficient manipulation of objects.
Establishing and breaking contact is especially hard because it causes a switch in the dynamics of the system which can rapidly lead to failures if not controlled properly. Unforeseen changes in the contacts location and properties (friction, stiffness, etc) can also dramatically degrade the robot behavior and remain a fundamental challenge in robotic manipulation and locomotion.

Deep reinforcement learning has shown a lot of promise in recent years for robotic applications. However, in an effort to learn end-to-end policies the focus has often been on the complexity in the observation part of the task, specifically vision, and not necessarily on the physical interaction part of the problem. One important aspect, that we investigate in the paper, relates to the choice of a policy parametrization that affords efficient learning of policies robust to contact uncertainties.

It was shown that position control with fixed pre-tuned gains can have better learning performance than pure torque outputs \cite{peng2017learning}, and such policies have been successfully transferred to physical robots interacting with the environment \cite{hwangbo2019learning}. These results are achieved by guiding properly the exploration using desired position and stabilizing the system around that using pre-tuned feedback gains.
However, previous work has demonstrated the importance of some form of force control when learning interaction tasks, either explicitly \cite{kalakrishnan2011learning}
or by learning time-varying control gains ~\cite{buchli2011learning,stulp2012model,kronander2013learning,luo2019reinforcement}, state-varying feedforward and feedback gains~\cite{gribovskaya2011motion}, or unified motion and gains varying strategy~\cite{khansari2017learning,viereck2018learning,martin2019variable}.

Recent results \cite{martin2019variable,8967946} suggest that learning impedance schedules in task space can significantly speed up learning of manipulation tasks.
The operational space representation used in those works has the advantage of abstracting the robot kinematics, but with the potential drawback of fixing the redundancy resolution scheme which can limit the range of possible behaviors. 
Indeed, the ability to vary nullspace resolution schemes is critical to enable the concurrent execution of several tasks necessary to achieve complex behaviors, e.g. avoiding an obstacle while reaching for an object or taking a step while maintaining balance \cite{saab2013dynamic, herzog2016momentum}.
One can also argue that methods that require solving an inverse problem suffer from numerical instability near singularity, or rule out a significant space of possible motions achievable without pre-defining a task-space.
Moreover, there is evidence that the best choice for a task-space may vary across and within tasks \cite{silverio2015learning}.
For these reasons, in this paper we focus on joint space policy learning despite potential training speedup that can be achieved by doing the same in a predefined task space.

The main goal of this paper is to investigate the effect of policy parametrization on reinforcement learning for robotic tasks involving complex contact interactions and hard impacts.
We provide empirical evidence that control policies concurrently generating
desired positions and joint impedance tend to produce more robust behaviors. We present both extensive numerical simulations and real hardware experiments.
In particular, we find that the resulting policies are robust to various types of contact uncertainties (friction, stiffness and contact location).
Additionally, we propose a reward term regularizing these variable gain policies and giving them interpretability, allowing for direct transfer to a real robot.
We perform an extensive analysis on two very different systems: a single-leg hopper (floating-base) creating intermittent contacts with hard impacts on the ground and a manipulator (fixed-base) performing a delicate force control task.
In both cases we show that variable gain control outperforms a wide range of learned fixed gain or direct torque control policies, 
especially in the presence of contact uncertainty.

\color{black}

\section{CONTROL POLICY PARAMETRIZATION}

\begin{figure*}
\includegraphics[width=0.8\linewidth]{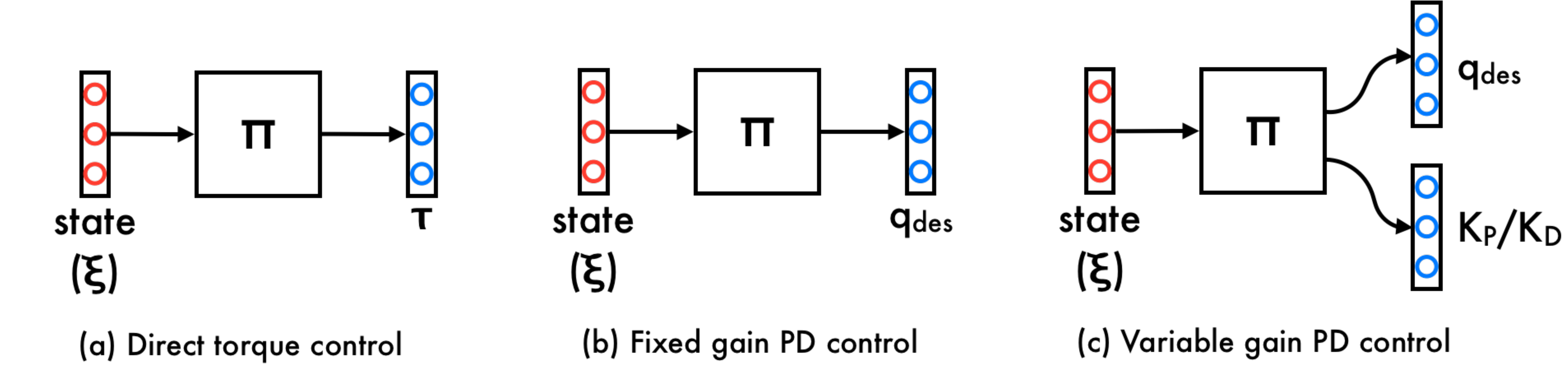}
\centering
\caption{Structures for the three control policies used.}
\label{fig:controller_param}
\end{figure*}

In this work, we compare several ways to parameterize control policies when using reinforcement learning in robotics tasks.
For each examined control parametrization we represent the policy with a neural network.
The inputs of the network stay the same in each case (the state of the system, $\bm{\xi}$), while the output corresponds to the parameters of the individual controller (torque, desired positions, gain parameters, etc).
Based on these parameters the controller produces the torque, $\bm{\tau}$, that is then applied to the system.
We now present each of the controller parametrization that will be examined.

\medskip

\textbf{Direct torque control.} In the first parametrization, the neural network directly outputs desired torque commands (Figure \ref{fig:controller_param}(a))
and
the resulting control law is simply:
\begin{equation}
    \bm{\tau} = \bm{\tau}(\bm{\xi})
\end{equation}
This parametrization imposes no structure, which provides some benefits.
As there is no imposed structure, any control that is a function of the variables present in the state can be expressed using this parametrization, given that the neural network is capable of approximating it.
Additionally, such control parametrization may have a more direct control of interaction forces than position controllers and may provide benefits in that respect, at least in static contact situations.

However, having no imposed structure also has several downsides.
With this parametrization, the function to be learned may be unnecessarily complex.
For example, for a motion task, the policy needs to learn something akin to PD control as well as a part that generates desired positions, all inside one function. This representation does not explicitly separate the feedback (how to correct from errors)
from the feedforward (what is the desired behavior) pathways.
Further, small variations in commanded torques can lead to very different movements and therefore to different task costs.
This will also raise issues for generating meaningful exploration, which might be more difficult than doing the same in the space of desired joint positions.

\medskip

\textbf{Fixed gain PD control.} The second parametrization considers a PD controller with fixed gains.
In this case, the policy outputs desired joint positions and torques are computed using a PD controller with some pre-defined gains (Figure \ref{fig:controller_param}(b)).
The control law is then
\begin{equation}
\bm{\tau} = \bm{K_p} (\bm{q_{des}}(\bm{\xi}) - \bm{q}) - \bm{K_d} \bm{\dot{q}}
\end{equation}
Again, there are positive and negative aspects to such a parametrization.
It is often easier to find solutions in this setup \cite{peng2017learning}, due to a simpler action space to explore and a policy easier to encode. Indeed, small variations in desired positions lead to small variations in task execution, at least in contact-free motions.
It is in fact, as we will also see in our experiments later, the best choice for tasks involving only free space motions.

On the other side, achieving a desired behavior in interaction with the environment becomes more difficult, especially when uncertainties are present, even when the gains are well tuned for the specific task.
The policy can to some extent control the interaction with the environment by changing desired joint positions, but as we will see later, finding such solutions becomes increasingly difficult.

\medskip

\textbf{Variable gain PD control.} The last parametrization is a PD controller with variable gains, i.e. 
the policy modulates both the desired position and impedance of each joint (Figure~\ref{fig:controller_param}(c)).
The control law is written as
\begin{equation}
\bm{\tau} = \bm{K_p}(\bm{\xi}) (\bm{q_{des}}(\bm{\xi}) - \bm{q}) - \bm{K_d}(\bm{\xi}) \bm{\dot{q}}
\end{equation}
\textit{Note:} In our experiments, we use a single output to control both $\bm{K_p}$ and $\bm{K_d}$ by imposing
a fixed relationship between damping and stiffness, similar to the scaling of critical damping.
The neural network outputs the $\bm{K_p}$ gain and $\bm{K_d}$ is varied with the square root of that value.

In contrast to the previous two controllers, we have introduced an extra degree of freedom (the gain modulation) for each joint. 
In theory, this added degree of freedom allows for an explicit separation between the feedforward path, i.e. the desired behavior encoded in the desired position, and the feedback path, i.e. the response to unforeseen events encoded in the feedback gain.

This parametrization will preserve the ease of exploration characteristic of PD control with fixed gains.
Moreover, with additional control dimensions to use, the functions the policy needs to learn may become even simpler than in the pure position control case.
Finally, key to contact sensitive tasks we are particularly interested in, the policy has finer control over contact interactions with the environment. Robustness to environment uncertainty might be easier to encode in the feedback path while preserving the feedforward one to encode the ideal, unperturbed, behavior.

An obvious drawback is that we doubled the size of the action space the policy is acting in. But, as we will see later, this rarely causes loss in learning performance.

\medskip

\textit{Remark:} Both fixed and variable gain PD control we discuss in this work are different than the ones commonly used in robotics, where the control follows a predefined time-based trajectory \cite{hwangbo2019learning}.
All controllers we examine are state-based without any notion of time, and as such are capable of handling uncertainty in contact location and time.

\medskip

\begin{figure}
    \centering
    \begin{subfigure}[b]{0.32\linewidth}
        \centering
        \includegraphics[width=\linewidth]{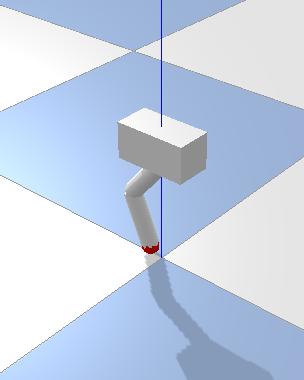}
        \caption{}
        \label{fig:B}
    \end{subfigure}
    \begin{subfigure}[b]{0.32\linewidth}
        \centering
        \includegraphics[width=\linewidth]{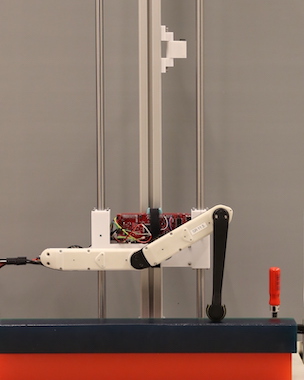}
        \caption{}
        \label{fig:A}
    \end{subfigure}
    \begin{subfigure}[b]{0.32\linewidth}
        \centering
        \includegraphics[width=\linewidth]{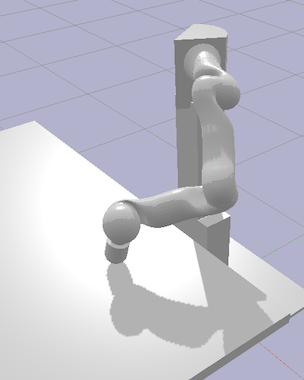}
        \caption{}
        \label{fig:A}
    \end{subfigure}
    \caption{Evaluation environments used: A floating-base system: a hopper jumping on a surface in simulation (a) and on real hardware (b); A fixed-base system: a manipulator interacting with the environment in simulation (c).}
    \label{fig:setups}
    \vspace{-5mm}
\end{figure}

\section{EVALUATION PROCEDURE}

\textbf{Evaluation goals.}
In our evaluations we focus on setups where proper interaction with the environment is crucial for task success.
In contact sensitive problems, planning and optimization-based approaches often struggle and reinforcement learning has the potential to generate solutions which cannot be easily found otherwise.

A central aspect of this work is to evaluate how controller parametrizations influence task performance  in presence of different contact uncertainties in the environment.
This is an important aspect in order to generate motions that can transfer to real physical systems.
To be able to successfully transfer policies from simulation to real systems we need to find solutions that are capable of handling variation in critical environment parameters.
Therefore, the parametrization best able to find solutions in such cases is more likely to produce good results when applied on the physical system.

\medskip

\textbf{Evaluation environments.}
To study these aspects, we use two different environments, a single leg hopper which is a floating-base system and a manipulator fixed to the ground.
We seek to show that our results are consistent across two very different environments, with contact interactions of completely different nature.
The manipulation task requires fine interaction while the hopping task requires
soft, reactive landing and quick force exertion. %
Both contain movements that alternate between free-space motions and contact interaction phases.
The simulations are implemented in the PyBullet simulator \cite{coumans2016pybullet}.
We also show direct transfer of the learned policy for the hopping environment to the real system.

\medskip

\textbf{Reinforcement learning algorithm.}
For training the control policies we use Deep Deterministic Policy Gradient (DDPG, \cite{DBLP:journals/corr/LillicrapHPHETS15}).
We chose an off-policy algorithm to reduce the issue of local minima, especially present here arising from combination of learning to control in joint space, discontinuities in the dynamics arising from contact interaction, and complex, multi-part reward functions.
However, we do not make use of any particularities of DDPG in our approach. We therefore expect that the results we present here will remain consistent when using other learning algorithms.

\section{CONTROLLING A FLOATING-BASE SYSTEM}

\subsection{Task description}

\textbf{Setup.}
The first setup we use in our evaluations consists of a floating-base robot hopper with a two degrees of freedom leg \cite{grimminger2020open} and a solid surface beneath it.
We restrict the base to only move along the Z-axis which eliminates the falling down effect while still capturing the base motion and intermittent contacts during continuous jumping.

\medskip

\textbf{Task.}
The task is to achieve stable periodic hopping motions.
We penalize hard impacts on the ground, as it is not something that would be acceptable on the real system.
We are interested in motions where the system smoothly lands and pushes off, without any discontinuities in its velocity.
To produce policies that are robust to contact switch uncertainty we randomly change ground surface height during episode execution in a range between \SI{-5}{\centi\meter} and \SI{5}{\centi\meter}. This corresponds to ground variations of 31\% of the total hopper height in the fully stretched configuration (\SI{32}{\centi\meter}).

We set the state of the system to consist of the joint positions and velocities for the two leg joints as well as position and velocity of the base.
We do not explicitly provide to the system any information about contact.

\medskip

\textbf{Reward function design.}
To generate hopping motions we intentionally keep the reward function as simple as possible.
The main part of the reward is based on the height of the robot base at every timestep, with an increase for values that cannot be reached without leaving the ground. This term, on its own, is enough to produce consistent hopping motions.
However, regardless of the controller design, policies trained on such a reward produce exactly the excessive impacts on the ground we are looking to avoid.
In order to prevent impact forces that can produce damage on the real system we penalize large forces applied to the robot.
Finally, to avoid high frequency control command we introduce a torque smoothness penalty.

Even though in this case we are dealing with a comparably simple system,  this reward design creates a challenging learning problem.
It is relatively easy for policies to get stuck in a local minima where the system is just held upright with its leg fully extended and not reach any hopping motion in their exploration.
The addition of the large force penalty makes the problem even more difficult as initial hopping motions found during training are bound to result in penalty for bad landings larger than the reward received for jumping.

\subsection{Simulation results}

\begin{figure}
\includegraphics[width=\linewidth]{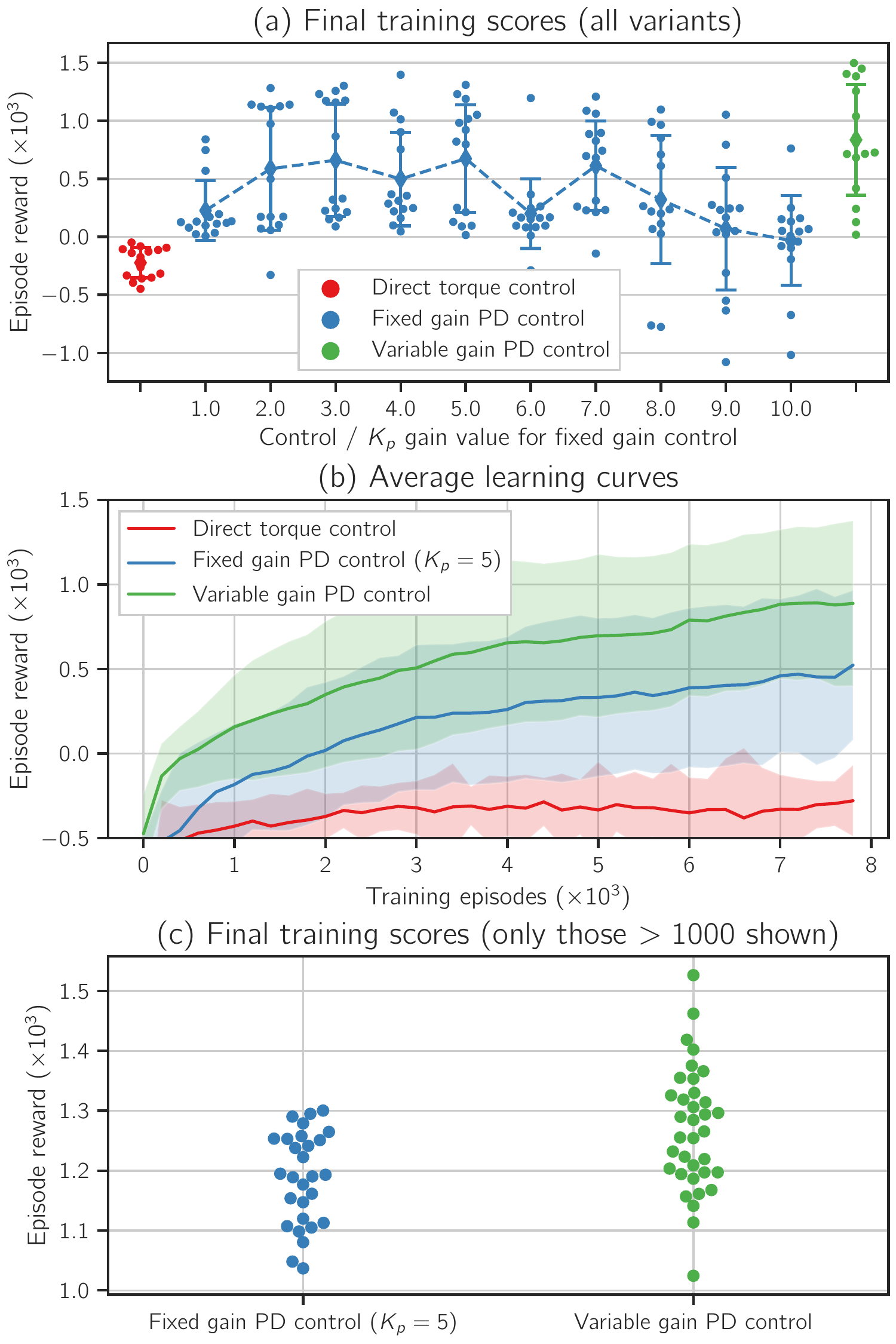}
\centering
\caption{Results for training in simulation on the hopping environment:
(a) Final training scores for all controller parametrizations.
For the fixed gain policy results are shown for a wide range of gain values;
(b) Averaged learning curves for the three controller parametrizations.
Results are averaged over 100 training instances for each parametrization;
(c) Closer look at the best performing policies at the end of the training (only those with score greater than 1000).
}
\label{fig:hopper-combined}
\vspace{-5mm}
\end{figure}
\textbf{Quantitative comparison.}
We run several training instances for each of the three controller parametrization.
For fixed gains control we examine performance over a range of different gains.
In Figure~\ref{fig:hopper-combined}(a) we show final scores at the end of each training.
We can see that direct torque control policies completely fail to achieve the task.
For fixed gain control we can observe a drop in performance when gains are too low and when they are too high, with a medium value of $K_p=5.0$ having the best average performance.

For reliably comparing fixed and variable gain control on this task we repeat the experiment using 100 training instances for each controller parametrization, using the best performing gain value for fixed gains control.
In Figure~\ref{fig:hopper-combined}(b) we present learning curves averaged over all the experiments for each controller parametrization.
We can see that average performance of variable gain control exceeds that of fixed gain one, a result of more of those policies converging to adequate solutions as well as doing so faster.

Being more likely to converge is important, as it allows us to find policies we can use in a limited number of training runs.
However, at the end we are always going to pick the best performing policy to deploy on the real system.
We are therefore interested in understanding how good the \textit{best policies} are.
In Figure~\ref{fig:hopper-combined}(c) we show the individual final performances from the previous experiment for all the policies above a score of 1000.
Here we can see that it is not only the case that the variable gain policies are more likely to converge or that they do so faster on average, but that they find solutions strictly better than any that are found with the fixed gain parametrization.

\medskip
\textbf{Qualitative analysis.}
\begin{figure*}
\includegraphics[width=\linewidth]{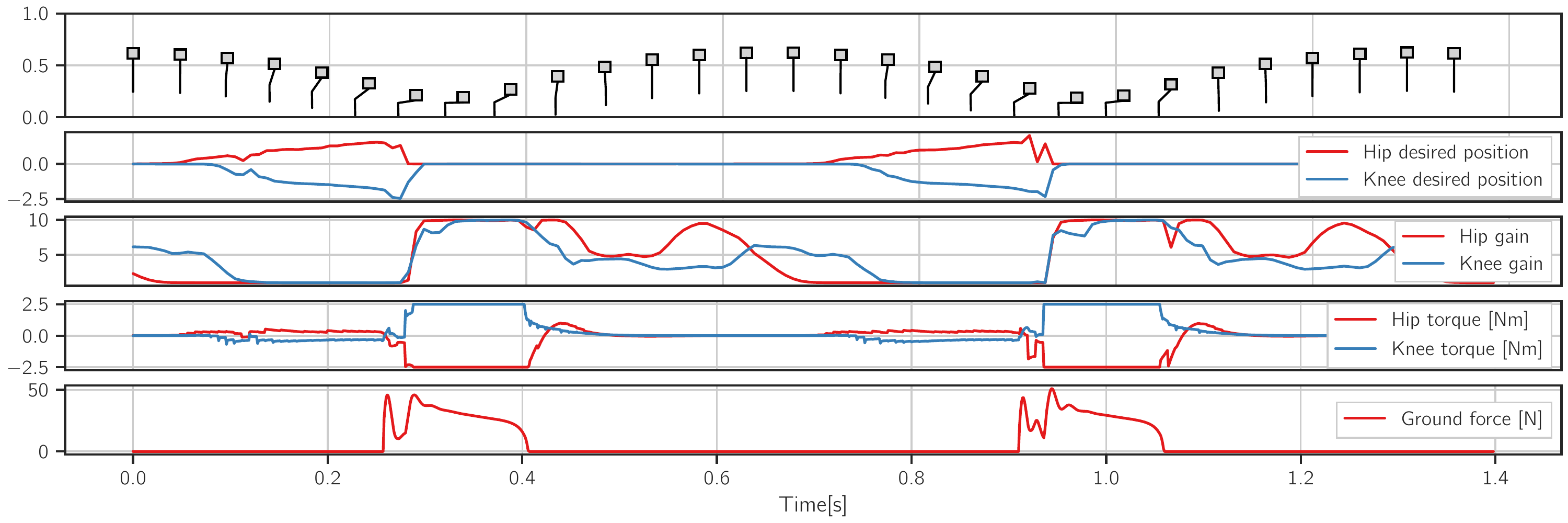}
\centering
\caption{Example of a learned behavior of the variable gain policy in simulation. The gains are at minimum just before landing and then increase quickly to enable push-off.}
\label{fig:hopper-qualitative}
\end{figure*}
In Figure~\ref{fig:hopper-qualitative} we show sample of behavior of the best performing variable gain policy.
Examining the gain profiles, we can note that they go to lowest possible value in advance of making contact, before spiking up to allow the system to first slow down, and then push itself off the ground.
The torque values stay close to zero except during landing and push-off when they go to maximum values that can be exerted.
The contact interaction is such that the impact is absorbed and the force is smoothly applied to first slow down and then accelerate the system upwards, without losing contact at any point.

\subsection{Experiments on the real system}

\textbf{Transferring policies to real systems.}
When learning in simulation the policies can learn to exploit idealized rigid-body dynamics.
Indeed, simulations do not include robot flexibility or difficult to model dynamic effects such
as Coulomb friction, drive train dynamics, etc. Furthermore, they typically assume infinite bandwidth
control authority, without any delays, which is known to be an issue to compute optimal policies \cite{grandia2019frequency, hwangbo2019learning}.
As a result, optimized policies can quickly change control outputs which results in very good behavior
in simulation but excites the dynamics of the real robot in problematic ways, for example creating
unwanted and possibly destabilizing oscillations in the motion.
This is an effect that we observed in our initial experiments.

An approach often taken in sim-to-real research is to randomize robot parameters in the simulation to 
capture various types of unmodeled dynamics \cite{peng2018sim}. However, there is no guarantee that this randomization
will capture the unmodeled dynamics of interest since the simulation does not explicitly capture this dynamics.
Further, dynamic randomization might prevent finding appropriate solutions that could transfer to the real robot by generating samples that do not apply to the real robot.
In this paper, we explore a different approach that exploits our control parametrization.%

\medskip
\textbf{Trajectory tracking reward term.}
\begin{figure*}
    \centering
    \begin{subfigure}[b]{\linewidth}
        \centering
        \includegraphics[width=\linewidth]{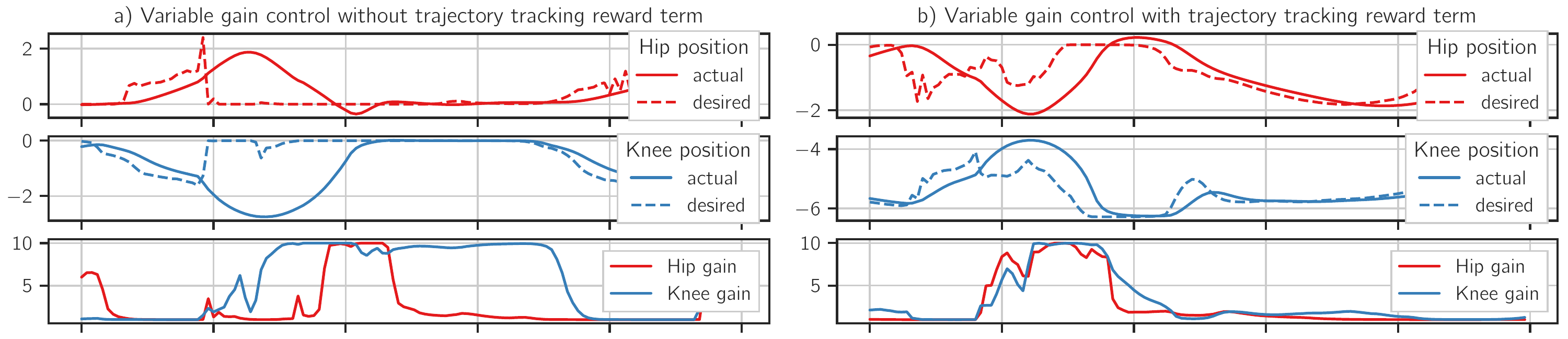}
        \label{fig:B}
        \vspace{-4mm}
    \end{subfigure}
    \begin{subfigure}[b]{\linewidth}
        \centering
        \includegraphics[width=\linewidth]{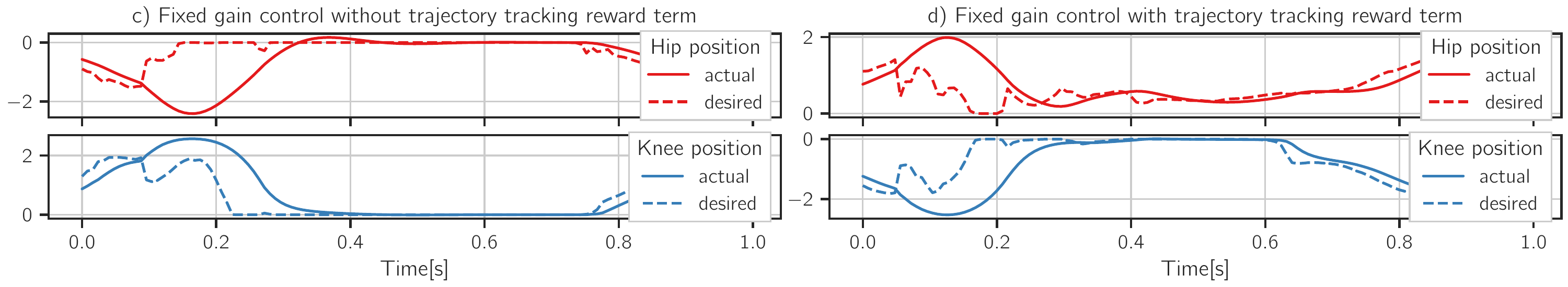}
        \label{fig:A}
    \end{subfigure}
    \vspace{-7mm}
    \caption{Effect of introduction of the trajectory tracking reward term on outputs of fixed and variable gain policies: (a), (b) Comparing the effect on resulting trajectory and gains for the variable gain policy; (c), (d) Comparing the effect on resulting trajectory in the case of fixed gain policy.}
    \label{fig:hopper-trajectory-tracking}
    \vspace{-5mm}
\end{figure*}
To address the aforementioned issue, we introduce an additional reward term during learning of the policies in simulation, which aims to force the policy to generate desired positions that can be effectively tracked.
The reward term penalizes the difference between the desired position given by the policy at time step $t$, $q_{des}^t$ and the actual position achieved in the next time step $t + 1$, $q^{t + 1}$ as follows:
$$
r_{tt} = -k \left\Vert q_{des}^t - q^{t + 1} \right\Vert^2
$$

With this reward term, we favor desired joint positions that can be tracked by the closed loop system in simulation.
Without such a reward term, the variable gain policy can make a choice between various desired position and gain pairs that results in the same applied torque at each time step.
For example, instead of generating desired positions that can be meaningfully executed, it can just give desired position values far from the actual robot trajectory with smaller gains to achieve the same torques and therefore the same behavior. However, this would lead to behaviors that could be very sensitive to variations in the dynamics.
As an additional benefit, this term forces the policy output to remain interpretable as desired position and feedback gains, clearly separating feedback and feedforward control paths.

We should note the difference between this term and a term penalizing desired positions that move away from the \textit{current} state of the system.
In the second case any desired position different than the current one receives some penalty and the system is incentivized not to move.
With the reward term we propose here if the position is reached in the next time step, zero penalty is given.
Even in the case when the desired position is not reached, penalty is only given on the remaining distance to it.
Only the desired values that cannot be reached are penalized, all motions where the trajectory can be tracked receive zero penalty.

In order to evaluate the effect of this addition to the reward on the resulting policies,
we repeat the training process described previously with the trajectory tracking penalty enabled for both fixed and variable gain policies.
We find that the best scoring policies for both these controller parametrizations still generate hopping with similar performance.

The outputs of both policies, with and without trajectory tracking reward, are shown in Figure~\ref{fig:hopper-trajectory-tracking}.
We can see that before adding this penalty term the variable gain policy gives desired position output far from the current trajectory when realizing the force needed to push off from the ground.
However, when training with this new reward term the desired positions for both joints track the actual trajectory more closely.
We can see that this also results in interpretable gain results -- going as low as possible before contact and then spiking up to realize the force that is needed.
On the other hand the output of the fixed gains policy practically does not change at all (the plots differ as a result of the opposite knee orientation found between the two solutions).
When gains are fixed, the only manner to control a desired contact force is to go off the trajectory with the desired position values. On the other hand, the resulting force can be controlled by varying the
gains in the other parametrization.

\medskip
\textbf{Evaluating policies on the real system.}
\begin{figure}
    \centering
    \begin{subfigure}[b]{0.18\linewidth}
        \centering
        \includegraphics[width=\linewidth]{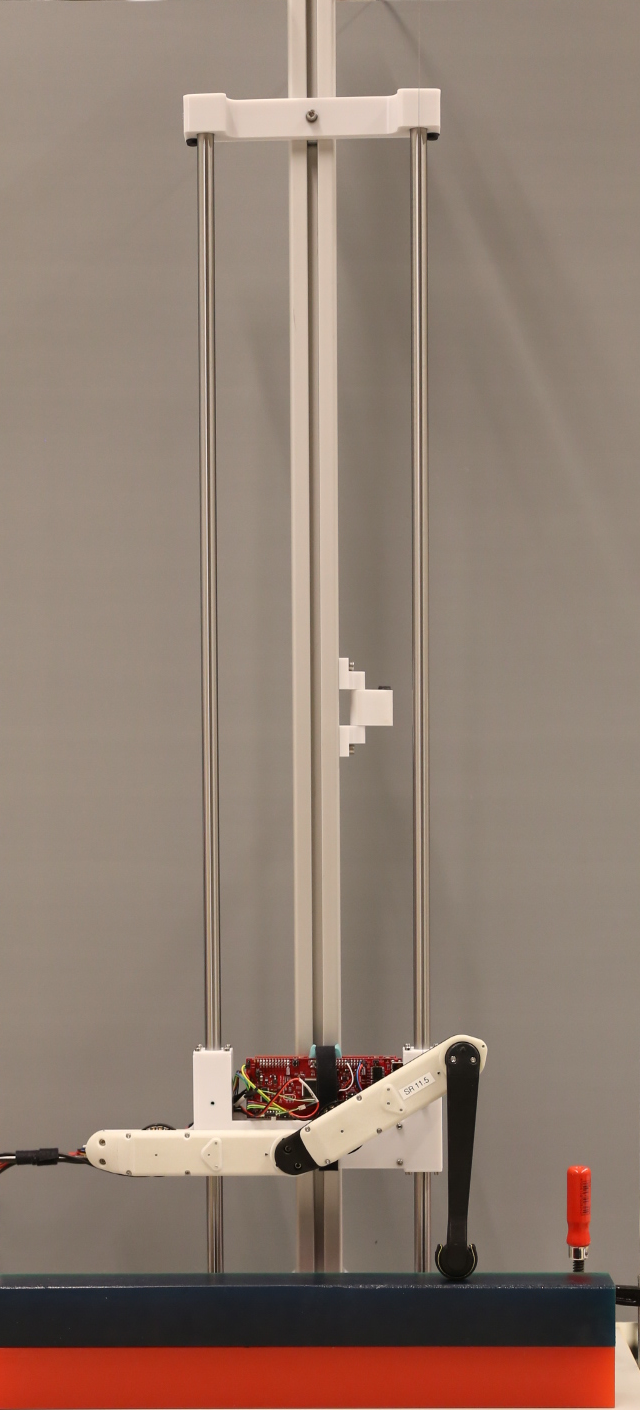}
        \caption{}
        \label{fig:A}
    \end{subfigure}
    \begin{subfigure}[b]{0.18\linewidth}
        \centering
        \includegraphics[width=\linewidth]{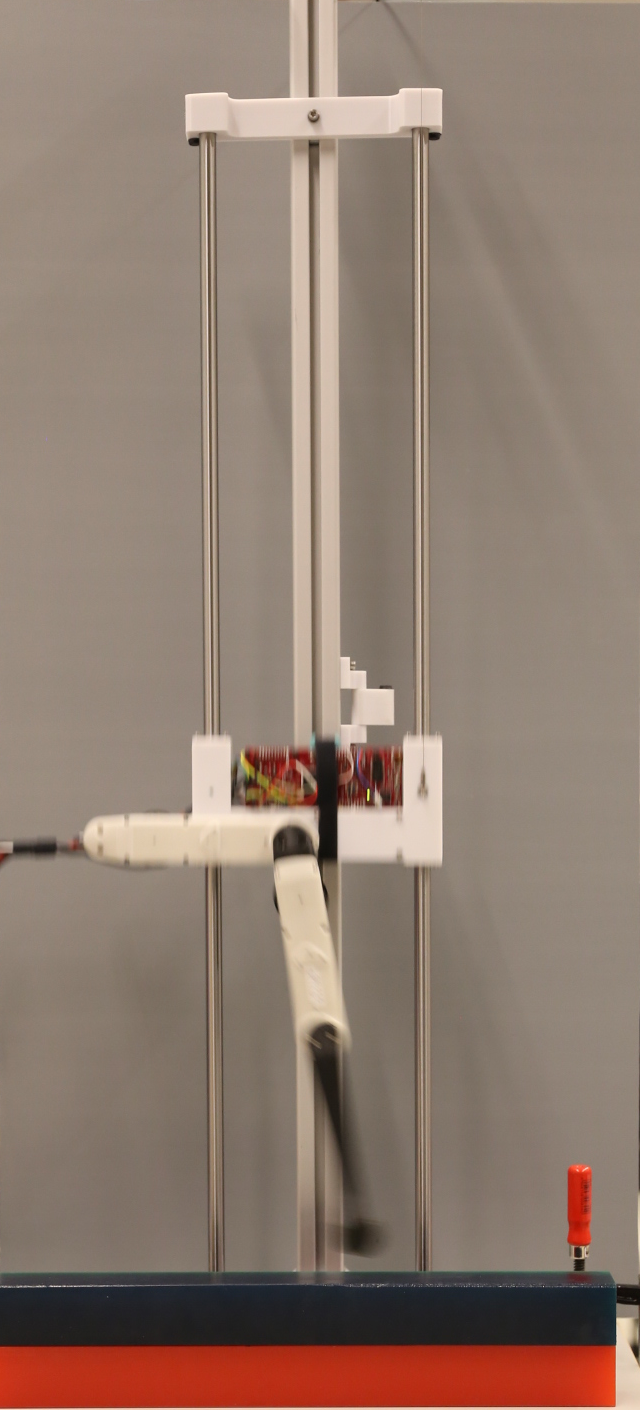}
        \caption{}
        \label{fig:B}
    \end{subfigure}
    \begin{subfigure}[b]{0.18\linewidth}
        \centering
        \includegraphics[width=\linewidth]{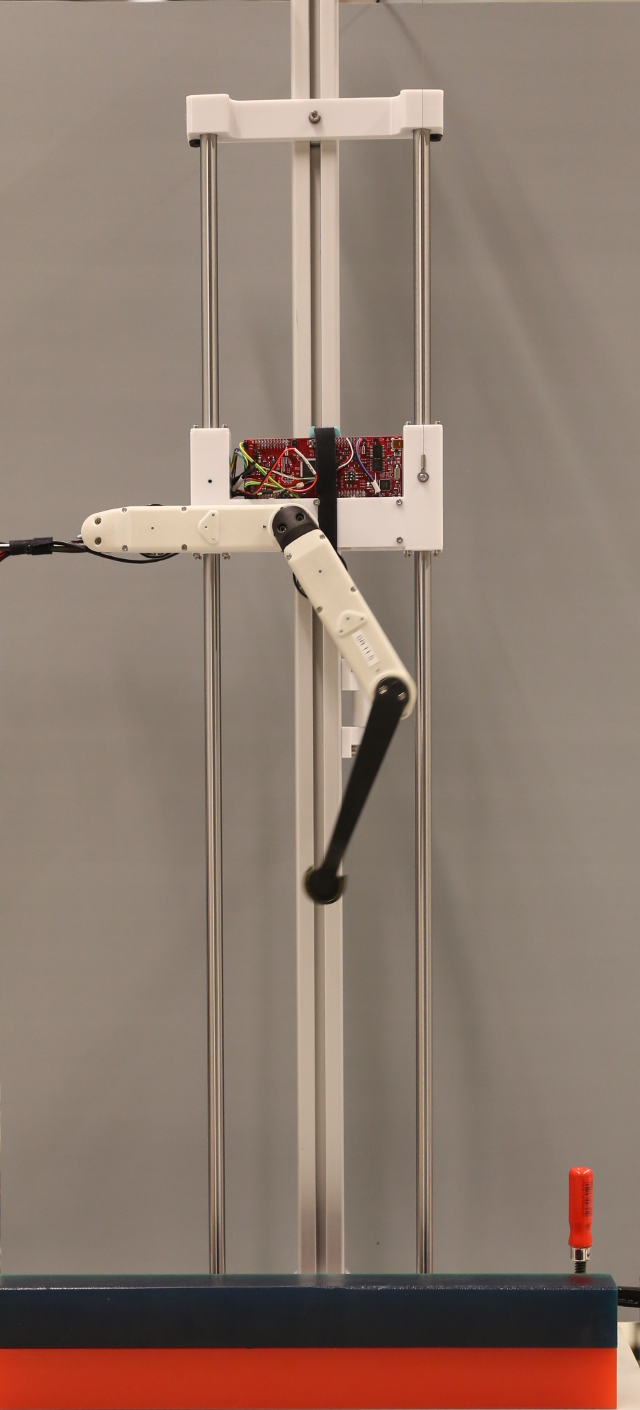}
        \caption{}
        \label{fig:A}
    \end{subfigure}
    \begin{subfigure}[b]{0.18\linewidth}
        \centering
        \includegraphics[width=\linewidth]{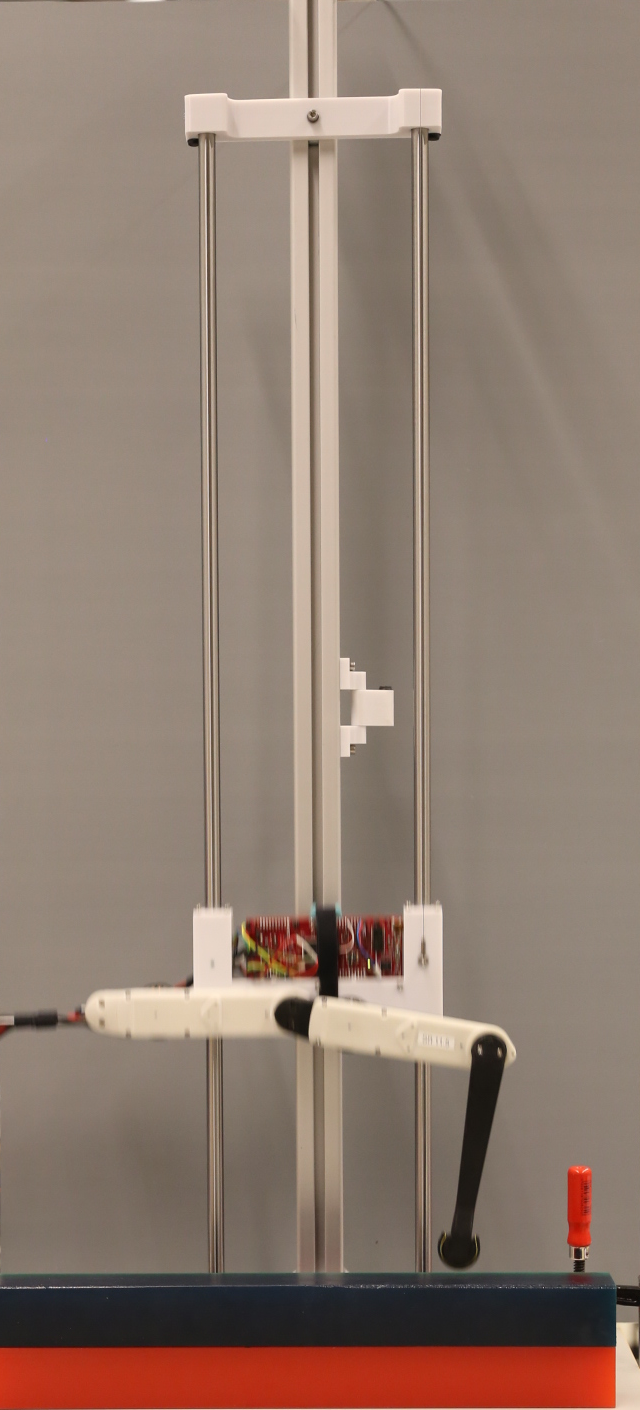}
        \caption{}
        \label{fig:B}
    \end{subfigure}
    \begin{subfigure}[b]{0.18\linewidth}
        \centering
        \includegraphics[width=\linewidth]{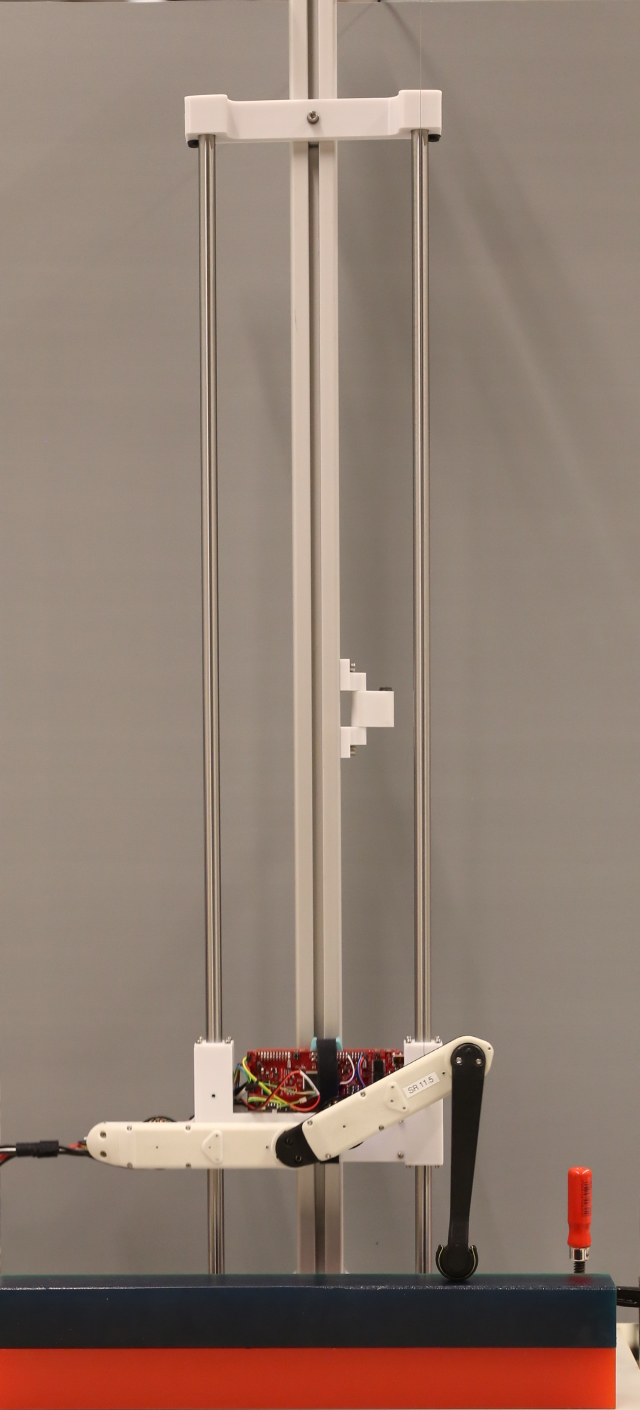}
        \caption{}
        \label{fig:A}
    \end{subfigure}
    \vspace{-3mm}
    \caption{Example of a variable gain policy behavior on the real system: (a) Push-off position on the ground; (b) Pushing off the ground; (c) Reaching maximum height; (d) Getting into position in preparation for the landing; (e) Landing and starting the cycle again.}
    \label{fig:photos-from-real}
    \vspace{-3mm}
\end{figure}
\begin{figure}
\includegraphics[width=\linewidth]{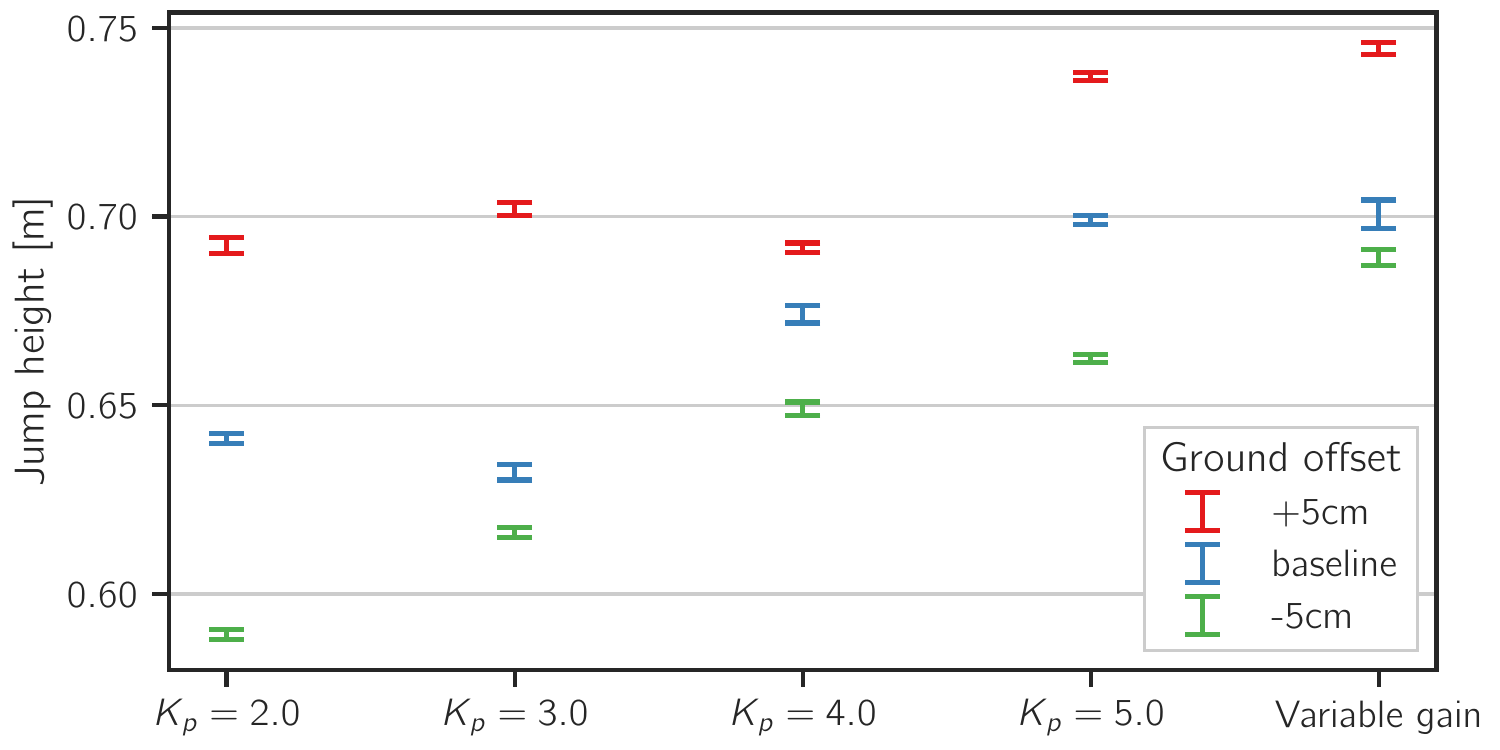}
\centering
\caption{Jumping heights for fixed and variable gain policies deployed on the real system. We use three ground positions where the baseline is the normal ground height and the other two (\SI{-5}{\centi\meter}, \SI{5}{\centi\meter}) are the ends of the range over which the policies are trained.}
\label{fig:hopper-real}
\vspace{-5mm}
\end{figure}
We evaluated the best performing policies in simulation for each gain value for the fixed gain controller and the one best performing variable gain policy.
Torque control did not produce any sensible policy worth evaluating on the real system.
For the fixed gain policies, having position gain values of $K_p=1.0$ was insufficient for hopping so that is omitted from the following comparison.
Gain values of $K_p=6.0$ and higher resulted in unstable behavior so those are omitted as well.

Apart from this, the transfer of policies trained in simulation in the way we described to the real system is very robust.
Out of the 6 policies we evaluated all succeeded in reasonable jumping performance, for many consecutive jumps, on the first try.
The achieved jump heights are presented in Figure~\ref{fig:hopper-real}.

Importantly, the variable gain policy outperforms all of the fixed gain ones, i.e. it is the policy
that saw the least decrease in performance.
Moreover, the resulting jumping height is higher than what was demonstrated
using model-based control in \cite{grimminger2020open}. It is in fact the highest jumping we 
were able to generate on this platform to date.
Also, even though its gains can, and do, go as high as $K_p=10.0$ it shows none of the instability present in the fixed gain policies with those same gain values.
The reason for this being that it increases the gains to maximum values only when necessary (e.g. during push-off). Indeed, the maximum admissible gains in contact will always be larger than during the flying phase due to the changing reflected inertia at the joints in loaded/unloaded conditions.

\color{black}
\section{CONTROLLING A FIXED-BASE SYSTEM}

\subsection{Task description}

\textbf{Setup.} We use a simulation of a 7 degree of freedom KUKA LWR manipulator (Figure \ref{fig:setups}(a)). In all the simulations, gravity is compensated with a feedforward term, which
is the default behavior on the real robot.

\medskip

\textbf{Task.} The task we are interested in consists of doing a circular motion with the endeffector touching a table in front of it, while applying a desired constant vertical force. This task is relevant for many applications that require sliding contacts, such as cleaning a surface, using a tool on an object, etc.
Tasks involving sliding contacts are especially difficult to optimize in general.
The task is designed such that the robot starts from a random initial position. It should be able to reach the table, establish a safe contact and exert a desired vertical force. We consider three types of uncertainties for the table: stiffness (i.e. how soft is the contact), friction, and height. These uncertainties are relevant for real-world applications as changes
in contact properties and location can easily destabilize controllers and  lead to failures. Moreover
contact stiffness and friction cannot be known precisely before interaction in an unknown environment.

For this task, the states of the system consist in the positions and velocities of all 7 joints of the arm, as well as the total force measured by the endeffector.

\medskip

\textbf{Reward function design.}  To learn a policy for achieving the desired task we define a reward function consisting of several parts (1-5). We use two terms to drive the circular motion along a desired trajectory: (1) the current distance from the endeffector to the closest point on the circle and (2) the difference between the current velocity vector and the desired tangential velocity on the closest point on the trajectory so as to achieve a motion with constant angular velocity. The three other terms are: (3) a reward based on the orientation of the endeffector, (4) a constant reward for any interaction between the endeffector and the table and a further reward based on the difference with desired contact force and (5) a penalty term for any interaction between the table and any part of the robot other than the endeffector.

While the reward function might seem complex, each term directly encodes one aspect of the task and we pay particular attention not to incentivize any specific behavior in solving it.
The apparent complexity is precisely the result of this, as we use multiple terms to define the circular motion along the trajectory, instead of simply having one specific point to track, explicitly to ensure time-invariance of the resulting policies.

\subsection{Simulation results}

\textbf{Quantitative comparison.}
We examine the robustness of our approach to variability present in the environment.
We consider in three separate simulations uncertainties on table height, friction, and stiffness.
For each of the three variables we define a possible range of values and uniformly sample a new environment in each episode during training.
We vary the table height in a \SI{20}{\centi\meter} range from \SI{0.8}{\meter} to \SI{1.0}{\meter} and Coulomb friction coefficients in a range from 0 (no friction) to 1.
We also vary the rigidity of the surface (which can easily influence the stability of a controller) with stiffness values from \SI{50}{N/m} to \SI{500}{N/m}.

\begin{figure*}
    \includegraphics[width=\textwidth]{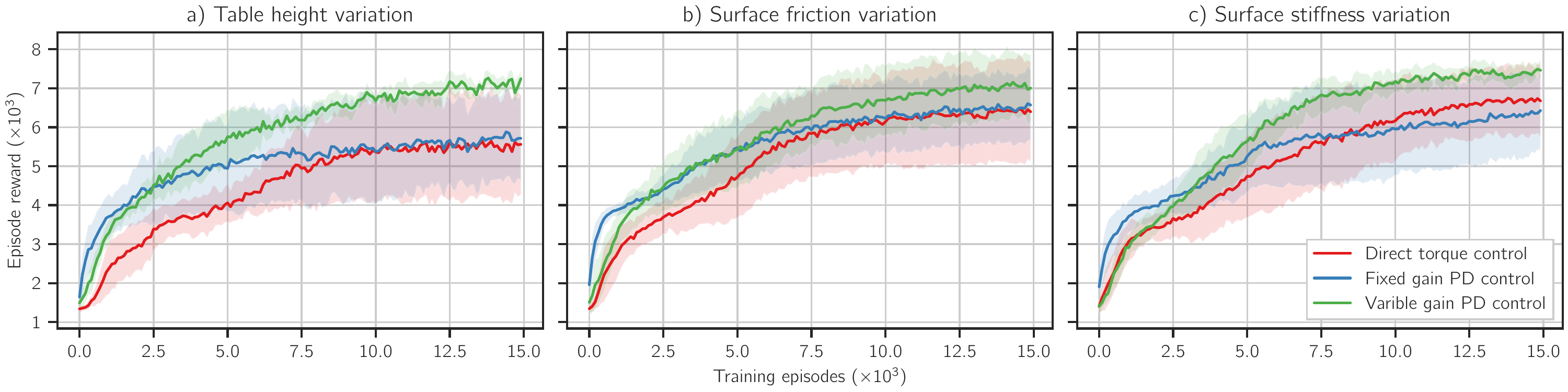}%
    \centering
    \caption{Robustness evaluations for the fixed-base setup. Results for environments with varying table height (a), table surface friction (b), and table surface stiffness (c).}
    \label{fig:apollo-robustness-combined}
    \vspace{-0.3cm}
\end{figure*}

We perform each policy training for a fixed, predefined number of episodes. For each controller we repeat the training 6 times, with different circular trajectories to track. In Figure~\ref{fig:apollo-robustness-combined} we present the combined results, showing the mean and standard deviation for the learning curves across these individual trainings.

We can see that variable gain control outperforms both of the other two parametrizations.
Splitting the control into motion and impedance parts makes it crucially easier for a good behavior to be found.
One control term can handle the circular motion, while the other, depending on the experiment, can manage contact location uncertainty or compensate for unknown friction of the surface.

\medskip

\begin{figure*}
    \includegraphics[width=\linewidth]{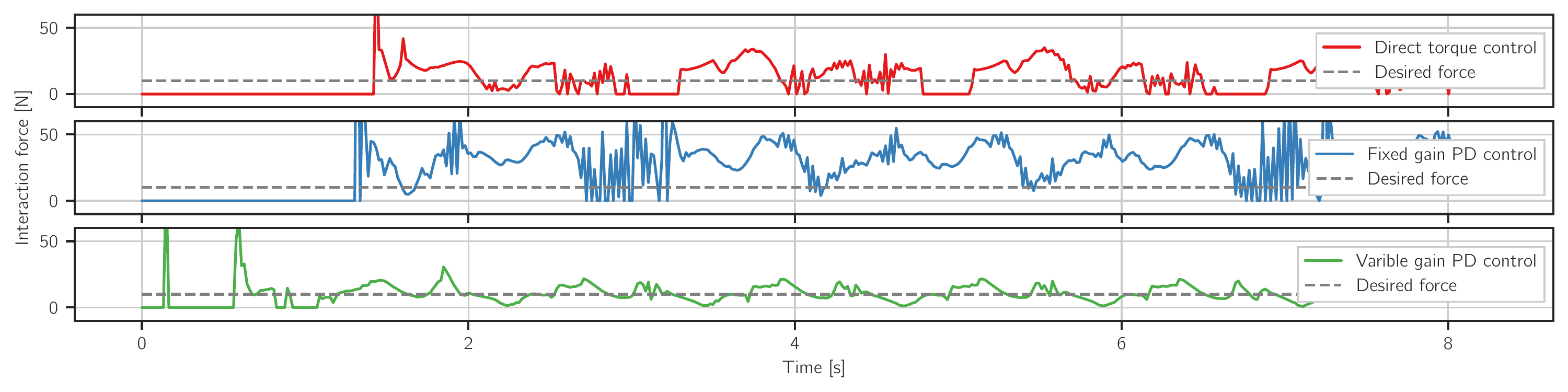}%
    \centering
    \vspace{-2mm}
    \caption{Table force interactions during circle tracking task for learned policies for different controllers.}
    \label{fig:apollo-qualitative}
    \vspace{-0.3cm}
\end{figure*}

\textbf{Qualitative analysis on contact transition.} Further comparison between different cases in Figure~\ref{fig:apollo-robustness-combined} reveals that the performance gap between the variable gain policies and the other two (fixed gains and direct torque) is more obvious when there is uncertainty in the contact location. Since the dynamics of the system changes before and after contact, transition between the two modes (i.e. free motion vs. in-contact) has a critical impact on the task achievement. Hence, the policy that is able to tolerate uncertainty in the mode transition can outperform other ones drastically. To investigate qualitatively the behaviour of different policies in this case, we plotted the corresponding normal interaction forces (Figure~\ref{fig:apollo-qualitative}) for a representative experiment. We can clearly see that the applied force from variable gain policy is smooth without loosing contact. On the other hand, direct torque control looses contact frequently and the fixed gains policy generates forces with high frequency oscillations. The variable gain policy leads to smoother contact forces which could be realistically applied on a real robot.

\section{Discussion}

\textbf{Trajectory tracking term.}
The trajectory tracking term was crucial in our sim-to-real transfer, as it prevents the policy from varying the desired position with a high frequency.
In other words, the policy is incentivized to change the desired position in a way that is consistent with the system dynamics and constraints (as much as it does not degrade achieving the desired task).
As a side benefit, the trajectory tracking term also gives interpretability to the output of the policy, i.e. the sequence of desired positions over time can be seen as the desired feasible trajectory and the multiplier to the error replicates the feedback gains.
As a result, if we have a variable gain policy, we can find a desired trajectory and an optimal set of feedback gains for that desired trajectory.
This interpretability can yield insight about the optimal impedance modulation for contact-rich tasks, which is still an open problem in the field.

\medskip

\textbf{Action space parametrization alternatives.}
When proposing any new structure in the action space of the policy, in addition to standard considerations on the control side, we suggest to take into account two additional factors from the learning perspective.
First, good exploration is critical for fast convergence and to avoid getting stuck in undesired local minima.
This is where position control based policies come ahead of direct torque control, but they in no way completely solve the problem and there is further research to pursue in that area.
Second, if we entirely decouple the feedback path from the feedforward one (e.g. $\tau = \tau_{ff}(\xi)-K_p(\xi)x - K_d(\xi)\dot x$), the learned policy may realize the entire control through a single term, mostly ignoring
the other one (our experiments with such control law formulations resulted in precisely that type of behavior).
This would also strip the control law terms of any physical meaning we tried to impose.
When using any such control law in a policy learning setting, where the same behavior can be produced by different combinations of control terms, there needs to be something incentivizing one choice over another.
Only then can the individual terms have the intended physical meaning, giving us the interpretability we desire.

\medskip

\textbf{Stability of variable impedance policy.}
Varying impedance as a function of state can cause instability of the control. As discussed in \cite{kronander2016stability}, reasonable varying stiffness profiles show no destabilization tendencies. In this paper, first we found a range of joint stiffness and damping that does not cause any instability on the real system for a wide range of motions. Then we let the policy find a state-dependent impedance within this range. With this strategy, we never experienced any instability when we applied the learned policy directly on the real robot.
\section{CONCLUSION}

In this paper, we have investigated the effect of action space representation on the performance and robustness in contact-rich tasks in the presence of uncertainties.
On both a floating-base hopping task and a fixed-base table wiping one we demonstrate that variable impedance control allows us to find better performing policies and to do so more reliably.
Additionally, we showed how we can use a regularization term to impose the original physical meaning to desired trajectory and impedance, giving interpretability to these policies.
Finally, we demonstrated how the policies can then directly be deployed on a real system, preserving performance and robustness.






\bibliography{example} 
\bibliographystyle{ieeetr}

\end{document}